\documentclass{article}
\usepackage{arxiv}
\usepackage[utf8]{inputenc} % allow utf-8 input
\usepackage[T1]{fontenc}    % use 8-bit T1 fonts
\usepackage{hyperref}       % hyperlinks
\usepackage{url}            % simple URL typesetting
\usepackage{booktabs}       % professional-quality tables
\usepackage{amsfonts}       % blackboard math symbols
\usepackage{nicefrac}       % compact symbols for 1/2, etc.
\usepackage{microtype}      % microtypography
\usepackage{lipsum}
\usepackage{graphicx}
\graphicspath{ {./images/} }
\usepackage{caption}
\usepackage{amsmath}
\usepackage{subcaption}
\usepackage{multirow}
\usepackage{rotating}

\title{A Novel Approach for Neuromorphic Vision Data Compression based on Deep Belief Network}

\author{
    Sally Khaidem\\
	Department of Electrical Engineering\\
    Indian Institute of Technology, Madras\\
    Chennai, 600036, India\\
    \texttt{ee20d041@smail.iitm.ac.in} \\
   \And
    Mansi Sharma\\
	Department of Electrical Engineering\\
    Indian Institute of Technology, Madras\\
    Chennai, 600036, India\\
  \texttt{mansisharma@ee.iitm.ac.in} \\
  \And
    Abhipraay Nevatia\\
	Department of Mechanical Engineering\\
    Indian Institute of Technology, Madras\\
    Chennai, 600036, India\\
  \texttt{me20b007@smail.iitm.ac.in } \\
}

\begin{document}
\maketitle
\begin{abstract}
A neuromorphic camera is an image sensor that emulates the human eyes capturing only changes in local brightness levels. They are widely known as event cameras, silicon retinas or dynamic vision sensors (DVS). DVS records asynchronous per-pixel brightness changes, resulting in a stream of events that encode the brightness change's time, location, and polarity. DVS consumes little power and can capture a wider dynamic range with no motion blur and higher temporal resolution than conventional frame-based cameras. Although this method of event capture results in a lower bit rate than traditional video capture, it is further compressible. This paper proposes a novel deep learning-based compression scheme for event data. Using a deep belief network (DBN), the high dimensional event data is reduced into a latent representation and later encoded using an entropy-based coding technique. The proposed scheme is among the first to incorporate deep learning for event compression. It achieves a high compression ratio while maintaining good reconstruction quality outperforming state-of-the-art event data coders and other lossless benchmark techniques. 
\end{abstract}

% keywords can be removed
\keywords{Event computing \and entropy coding \and dynamic vision sensor \and deep belief network}

\section{Introduction}
\label{sec:intro}
Sight, along with the brain, is the dominant sense in humans for perceiving the world and learning new things. “Silicon Retina”~\cite{mahowald1994silicon} mimics the neural architecture of human eyes and reveals a new, powerful way of computations, sparking the emerging field of neuromorphic engineering. Bio-inspired novel sensors such as Dynamic Vision Sensors (DVS)~\cite{lichtsteiner2008128} measure intensity changes asynchronously rather than capturing intensity images at a fixed rate. As a result, it generates a stream of events that encodes the time, location, and polarity of brightness changes, where the data rate depends on scene complexity and camera speed. When compared to traditional cameras, DVS have superior properties. They have a very high dynamic range (140 dB versus 60 dB), no motion blur, and measurements with latency on the order of microseconds. DVS devices, such as Dynamic and Active-pixel Vision Sensor (DAVIS)~\cite{brandli2014240} and Asynchronous Time-based Image Sensor (ATIS)~\cite{posch2010live} are a viable alternative in challenging conditions for standard cameras, such as high-speed high-dynamic-range motion photography, robotic automation, and intelligent surveillance~\cite{kim2008simultaneous,vidal2018ultimate,rebecq2019events,zhu2019unsupervised}.

The neuromorphic silicon technology uses Address-Event-Representation (AER)~\cite{chan2007aer}, a communication protocol for transferring spikes events between bio-inspired chips. A tuple $(X, Y, p, t)$ represents each event, where $X$ and $Y$ denote the location of the event at a particular timestamp $t$ with polarity $p$ indicating an increase or decrease in event brightness. Each tuple is represented by 64 bits, with the timestamp being 32 bits and the remaining three fields being 32 bits. The goal is to gather helpful information from event data and utilize it for processing.

\begin{figure}[!t]
     \centering
     \begin{subfigure}{0.35\textwidth}
    \centerline{\includegraphics[width=\textwidth]{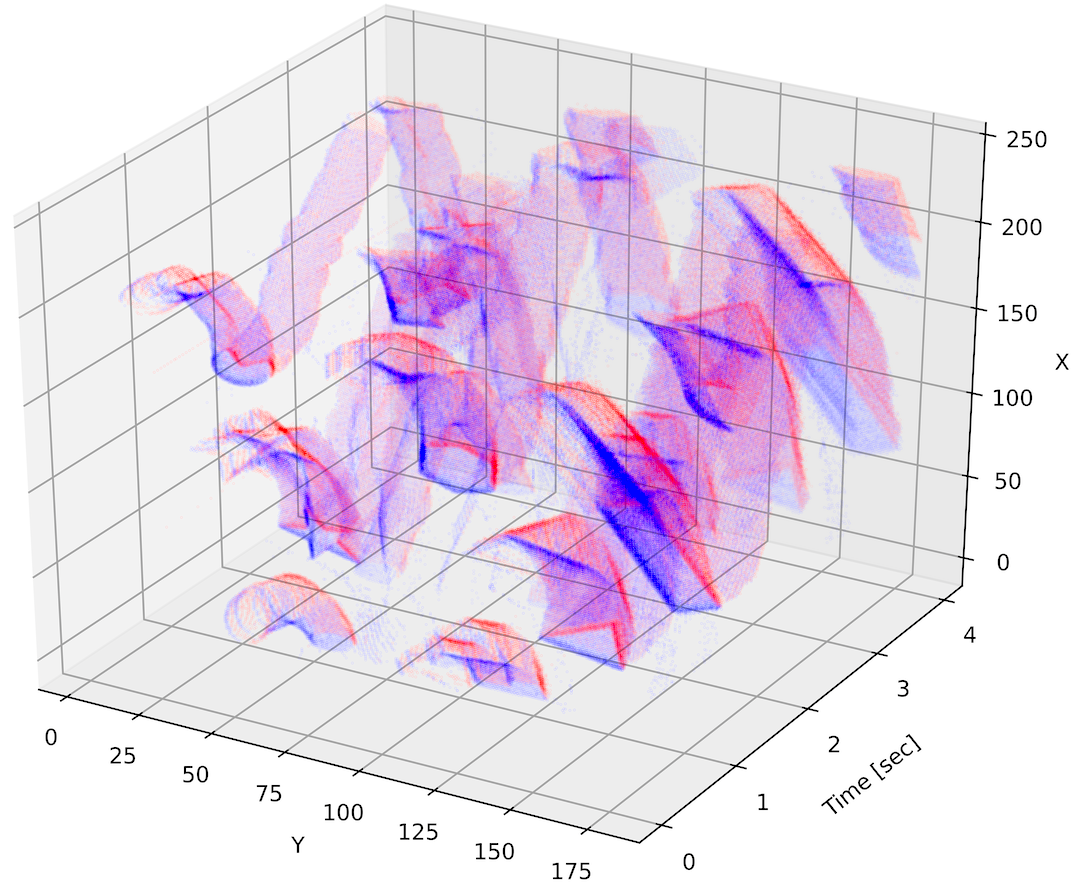}}
    \caption{}
    % \caption{Visualization of the event output in space-time without polarity separation. Red and Blue represents polarity $`0'$ and $'1'$ respectively.}
    \label{all_event}
     \end{subfigure}
     \begin{subfigure}{0.35\textwidth}
    \centerline{\includegraphics[width=\textwidth]{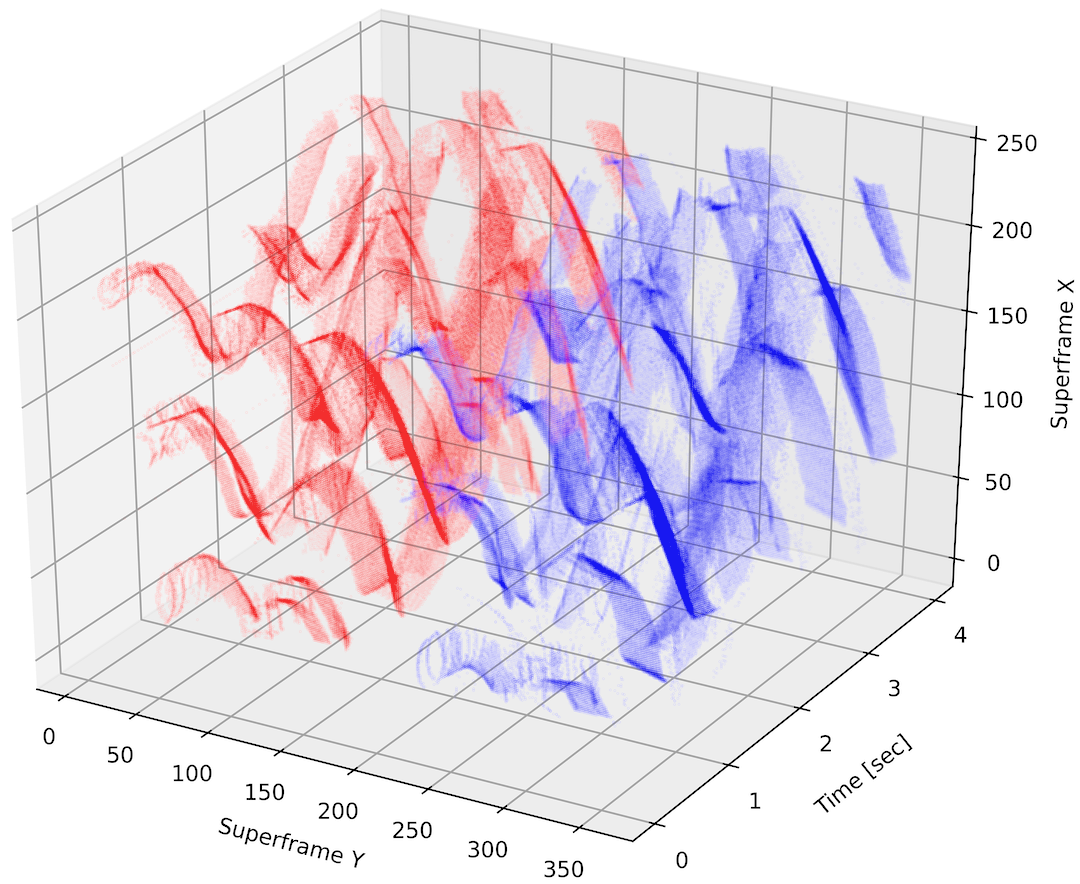}}
    \caption{}
    % \caption{Visualization of the event output in space-time with polarity separation. Red and Blue represents polarity $`0'$ and $'1'$ respectively.}
    \label{superframe}
     \end{subfigure}
    \caption{Visualization of \textit{Box} event output in space-time. Red and Blue represents events with polarity `$0$' and `$1$' respectively. (a) without polarity separation (b) with polarity separation and creation of super-frame.}
\end{figure}

DVS acquire information asynchronously and sparsely, with high temporal resolution and low latency. Hence, the temporal aspect, particularly latency, is critical in the event data processing. The output stream cannot use traditional vision algorithms since it is a series of asynchronous events rather than actual intensity images. Therefore, development of new algorithms that take advantage of the sensor's high temporal resolution and asynchronous nature is necessary. There are two types of algorithms based on the processing number of events at the same time. The first approach operates on an event-by-event basis, in which the system's state changes upon the occurrence of a single event, resulting in minimal latency. The second approach involves latency because it operates on groups or packets of events. It can still provide a system state update upon the occurrence of each event if the window moves by one event at a time. The data storage and transmission bandwidth limitation for onboard DVS processing is an open challenge and requires immediate solutions. Spike coding~\cite{bi2018spike} is a dedicated lossless compression strategy that exploits event data's time-series and asynchronous nature. It follows a cube-based coding framework where the spike sequence is divided into multiple macro-cubes and encoded accordingly. Entropy-based coding strategies like Huffman and Arithmetic can effectively encode DVS data by treating each spike event field as an input symbol. Existing lossless coding schemes such as dictionary-based~\cite{collet2018zstandard,deutsch1996zlib,alakuijala2016brotli} and fast-integer~\cite{blalock2018sprintz,lemire2015decoding} encoders can also compress the DVS data after converting the spike events into a multivariate stream of integers.

\begin{figure}[!t]
\centerline{\includegraphics[width=0.8\textwidth]{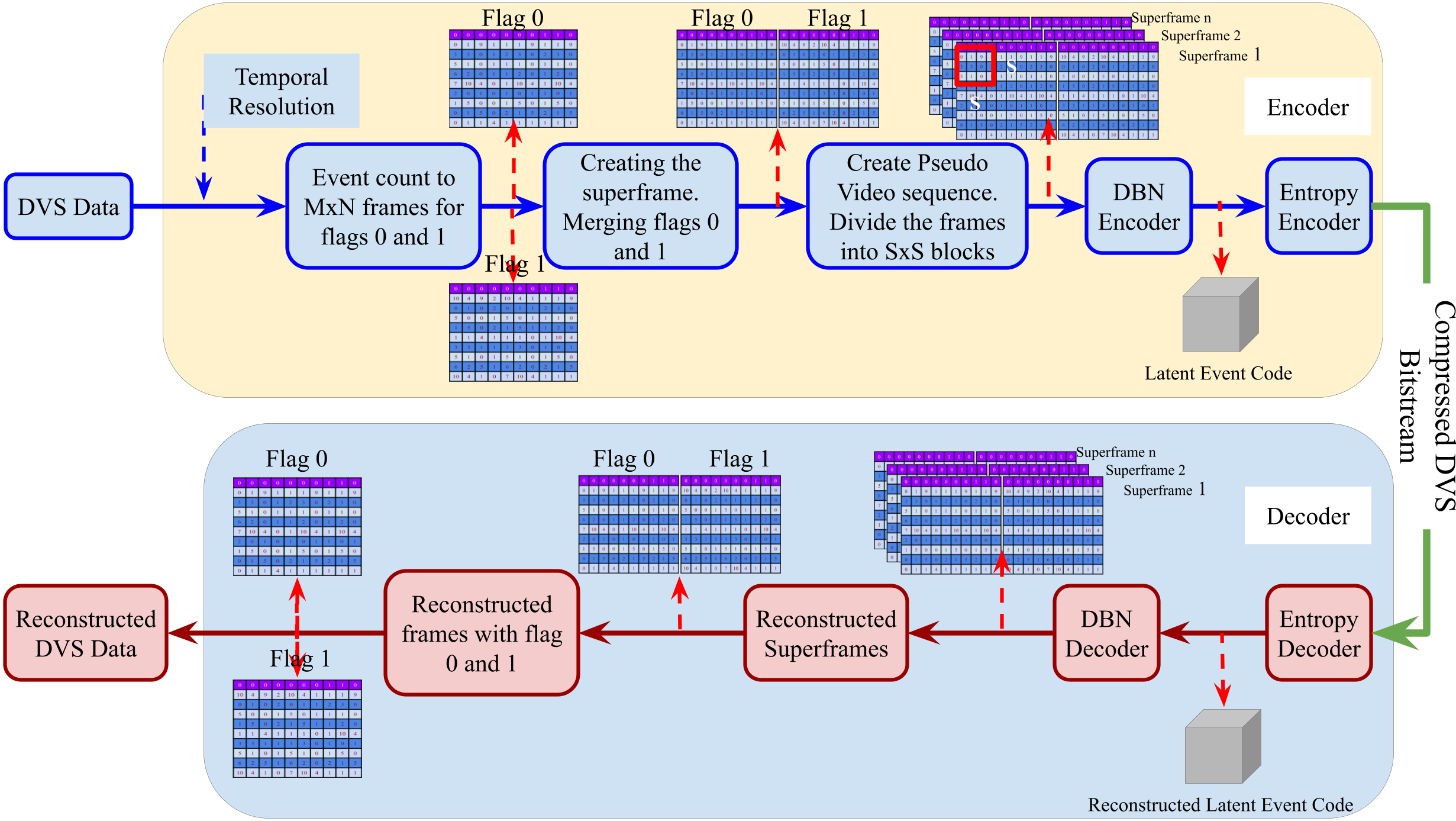}}
\caption{Complete workflow of proposed coding scheme.}
\label{Pipeline}
\end{figure}

The applications of DVS range from self-driving cars~\cite{maqueda2018event} to robotics~\cite{rigi2018novel} and drones~\cite{mueggler2014event}. Applications such as coordinating multiple intelligent vehicles (IoV) (cars, drones, etc.) having onboard processing constraints require real-time data sharing and feedback. In comparison to traditional sensing techniques, neuromorphic sensing provides an intrinsic compression. Further compression of event data is advantageous for transmission in the Internet of Things (IoT) and the Internet of IoV. This paper presents a novel approach suitable for DVS data compression based on a deep learning algorithm, Deep Belief Network (DBN). Figure~\ref{Pipeline} depicts the complete workflow of event compression. The entire stream of events is converted into a dimensionally reduced latent representation by multiple code layer blocks using the DBN. The compact latent code blocks contain recurring information suitable for lossless symbol-based encoders. Hence, we compress the latent code using an entropy-based Huffman coding technique. The primary contributions of the proposed scheme are as follows:
\begin{itemize}
    \item The proposed framework is among the first to incorporate deep learning techniques for event data processing. High-dimensional event data is transformed into low-dimensional latent code using a multilayer neural network called a deep belief network. We perform lossless encoding of low-dimensional latent features using entropy-based encoders to achieve a further compressed representation.
    \item We formulated a unique events arrangement deemed more suitable for processing by the proposed framework. The events are time-aggregated by accumulating spike events over time as super-frame sequences, as explained in Section~\ref{event_arrangement}. Super-frames result in high spatial and temporal correlation among the event data. 
    \item We conducted extensive comparisons with lossless benchmark strategies on a diverse standard dataset with varying scene complexity and camera movement. As a result of the learning-based framework, we obtain a concise latent code of high dimensional event data, increasing the compression gain while maintaining good reconstruction quality. Hence, the proposed framework outperforms the benchmark strategies.   
\end{itemize}

\section{Proposed Architecture}
\label{proposed_work}
DVS generates a high-dimensional multivariate data stream indicating each event's occurring timestamp, location and polarity. The proposed methodology's primary goal is to obtain an efficient representation of the high-dimensional input event data. Algorithms like auto-encoder encodes the input data into a much lower dimensional representation and store latent information about the input data distribution. Autoencoders typically find poor local minima with large initial weights; with small initial weights, the gradients in the early layers are tiny, making training autoencoders with many hidden layers impossible. In DBN, multiple RBMs obtain the initial weights for the autoencoder network. DBN serves as the foundation for the coding framework and reduces the high-dimensional event data into a compressed form. The latent version of events is a string of repetitive integer values that are further compressible using an entropy-based encoder such as the Huffman encoder.

\subsection{Event Arrangement}
\label{event_arrangement}
Even though the coding framework can encode DVS data on an event-by-event basis, we devise a unique arrangement of event data by applying time aggregation at particular time intervals. The primary benefit of event aggregation is the generation of temporal frames which reduces data size by projecting the DVS spike event stream into a series of frames recording the location histogram count, i.e., the number of event counts at each pixel. We segregate the spike sequence into two separate frames, one for increasing luminance (polarity 1) and one for decreasing luminance (polarity 0). On a particular pixel, if the previous event's polarity was zero (or one), the next event has a high probability of having zero (or one) polarity due to smooth change in luminance~\cite{bi2018spike}. The polarities of the accumulated spike events have a strong correlation so we record the event count separately for each polarity with each frame having full pixel array resolution. This increases the temporal correlation between frames of the same polarity~\cite{lagorce2016hots}. We combine the frames with the same timestamp from each polarity into a single super-frame as shown in Figure~\ref{Pipeline}. It comprises a `0' polarity frame on the left and `1' polarity frame on the right resulting in high inter-frame correlation. Hence, super-frames have inherent polarity information and reduce the required frame rate for processing, saving space to store the data. 
\subsection{Super-frames to Latent Representation}
Exploiting spatial, temporal, and statistical correlations among the event frames associated with the DVS sequence, we compress the accumulated spike events into latent code representation using a DBN. DBN is a stack of two-layered stochastic network with visible and hidden layers where the hidden layer of one RBM is the visible layer of next RBM. In binary RBMs, the random variables take the value $(v,h)\in \left\{0,1\right\}^{m+n}$ and the Gibbs distribution describes model's joint probability distribution $p(v,h)=\frac{1}{Z}e^{-E(v,h)}$ with the energy function $E(v,h)$.
\begin{equation}
\begin{aligned}
    E(v,h)=-\sum_{i=1}^{m}\sum_{j=1}^{n}w_{ij}h_iv_j-\sum_{j=1}^{n}b_jv_j-\sum_{i=1}^{m}c_ih_i
    \label{eq:energy_rbm}
\end{aligned}
\end{equation}
where, $w_{ij}$ is the weight associated with the nodes $v_i$ and $h_j$. The bias terms for $j^{th}$ visible and $i^{th}$ hidden node are $b_j$ and $c_i$  respectively. 
The change in weight is given as
\begin{equation}
\begin{aligned}
    \Delta w_{i j} = \epsilon (\left< v_i h_j \right>_{input} - \left< v_i h_j \right>_{recon})
    \label{eq:weight_rbm}
\end{aligned}
\end{equation}
where, $\epsilon$ is the learning rate and $\left< v_i h_j \right>_{input}$, $\left< v_i h_j \right>_{recon}$ denote the fraction of time event $i$ and feature detector $j$ are `ON' at the same time while driven by input and reconstructed output respectively.
\begin{figure}[!t]
     \centering
     \begin{subfigure}{0.26\textwidth}
        \centerline{\includegraphics[width=.7\textwidth]{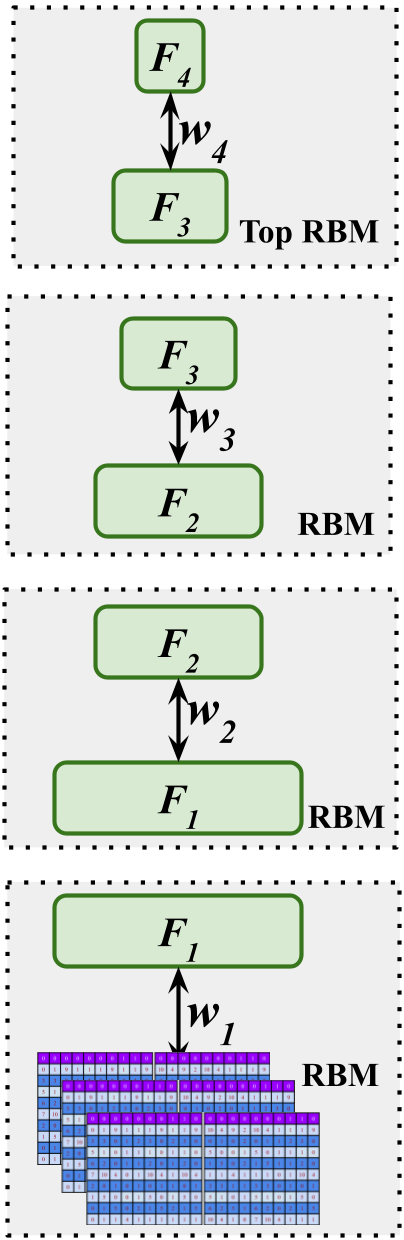}}
        \caption{}
        \label{pretraining}
     \end{subfigure}
     \begin{subfigure}{0.27\textwidth}
        \centerline{\includegraphics[width=.7\textwidth]{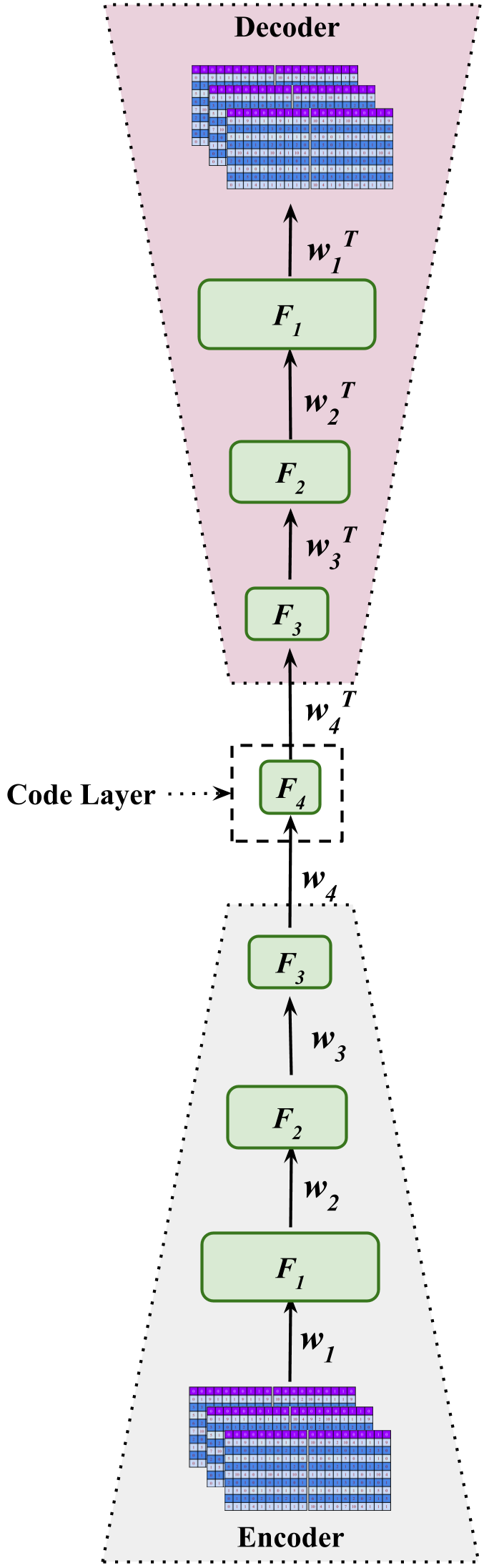}}
        \caption{}
        \label{unrolling}
     \end{subfigure}
     \begin{subfigure}{0.27\textwidth}
        \centerline{\includegraphics[width=.7\textwidth]{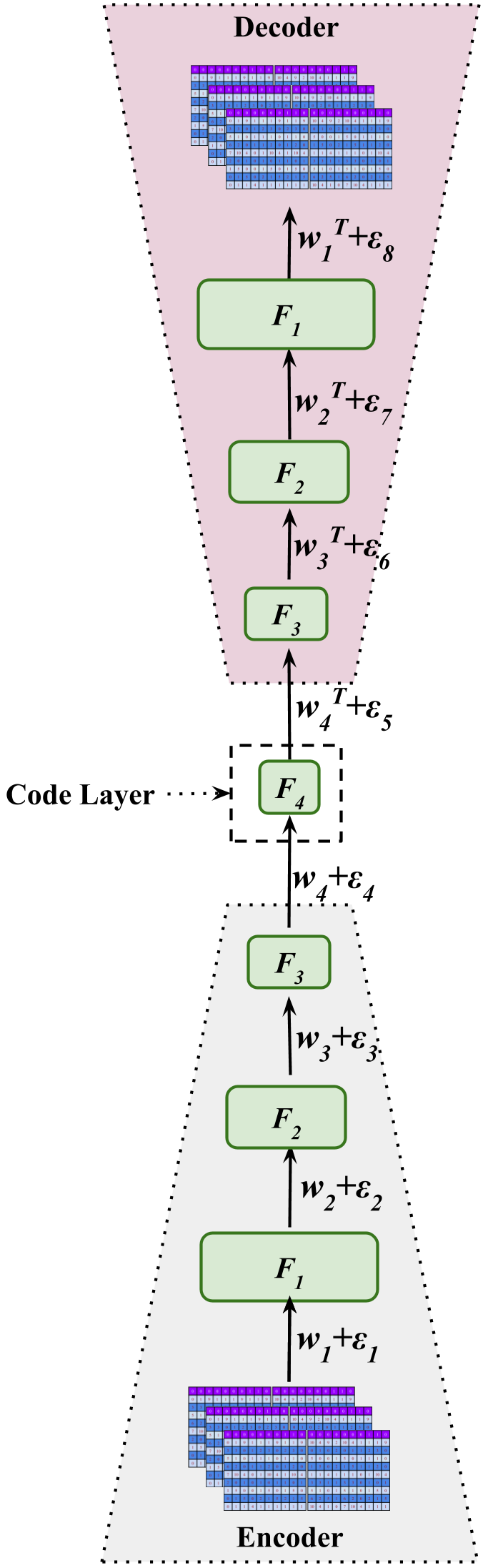}}
        \caption{}
        \label{finetuning}
     \end{subfigure}
        \caption{Workflow of deep auto-encoder with RBM pretraining.$(a)$ Pre-training of RBM weights $(b)$ Unrolling of stacked RBMs $(c)$ Fine-tuning of the unrolled stacked RBMs network.}
        \label{DBN}
\end{figure}

The layer-by-layer learning algorithm is a very effective way to pre-train the weights of a deep auto-encoder. Each feature layer captures high-order correlations between the events in the layer below. This gradually reveals the low-dimensional, nonlinear structures in a wide range of data sets. The workflow of DBN starts with pretraining process of feature detector layers with initial weights update~\eqref{eq:weight_rbm}, as shown in Figure~\ref{pretraining}. After pretraining is completed, the model is unfolded (Figure~\ref{unrolling}) to form a deep auto-encoder network initialised with the pre-trained weights. The pre-trained weights of deep auto-encoder are fine-tuned by replacing stochastic activities with deterministic, real-valued probabilities using the back-propagation method (Figure~\ref{finetuning}). For fine-tuning, we utilised the conjugate gradient method. The number of features extractors $F_1$, $F_2$, $F_3$ and $F_4$ in each layers are varied according to the requirement of input data such that $ F_2 > F_3 > F_4$ and $F_2 > F_1$ . The code layer with $F_4$ features is the latent space representation of the event super-frames. The latent code of accumulated spike events super-frame is a string of integer values. We used Huffman coder, an entropy-based encoder, to compress the latent code further. Huffman coding aims to reduce the expected code length by assigning shorter codes to frequently used characters and longer ones to rarely used characters. Adding Huffman coding to the DBN framework improves compression performance as discussed in section~\ref{experimental_results}, even though Huffman coding alone produces low compression gains. The encoder and decoder network share a symmetrical structure and number of features in each layer. The implementation details and experimental results are analysed in the later sections. 

\section{Experiments}
\label{experiments}
In this work, we first evaluate the proposed compression scheme's performance based on end-to-end and input-output compression ratios. The end-to-end compression ratio is between uncompressed events and compressed event bitstream size. In contrast, the input-output compression ratio is the input and output frame size ratio. Subsequently, we provide an extensive performance comparison with benchmark strategies regarding compression gains and peak-signal-to-noise ratio (PSNR). 
\begin{figure}[!t]
     \centering
     \begin{subfigure}{0.31\textwidth}
    \centerline{\includegraphics[width=\textwidth]{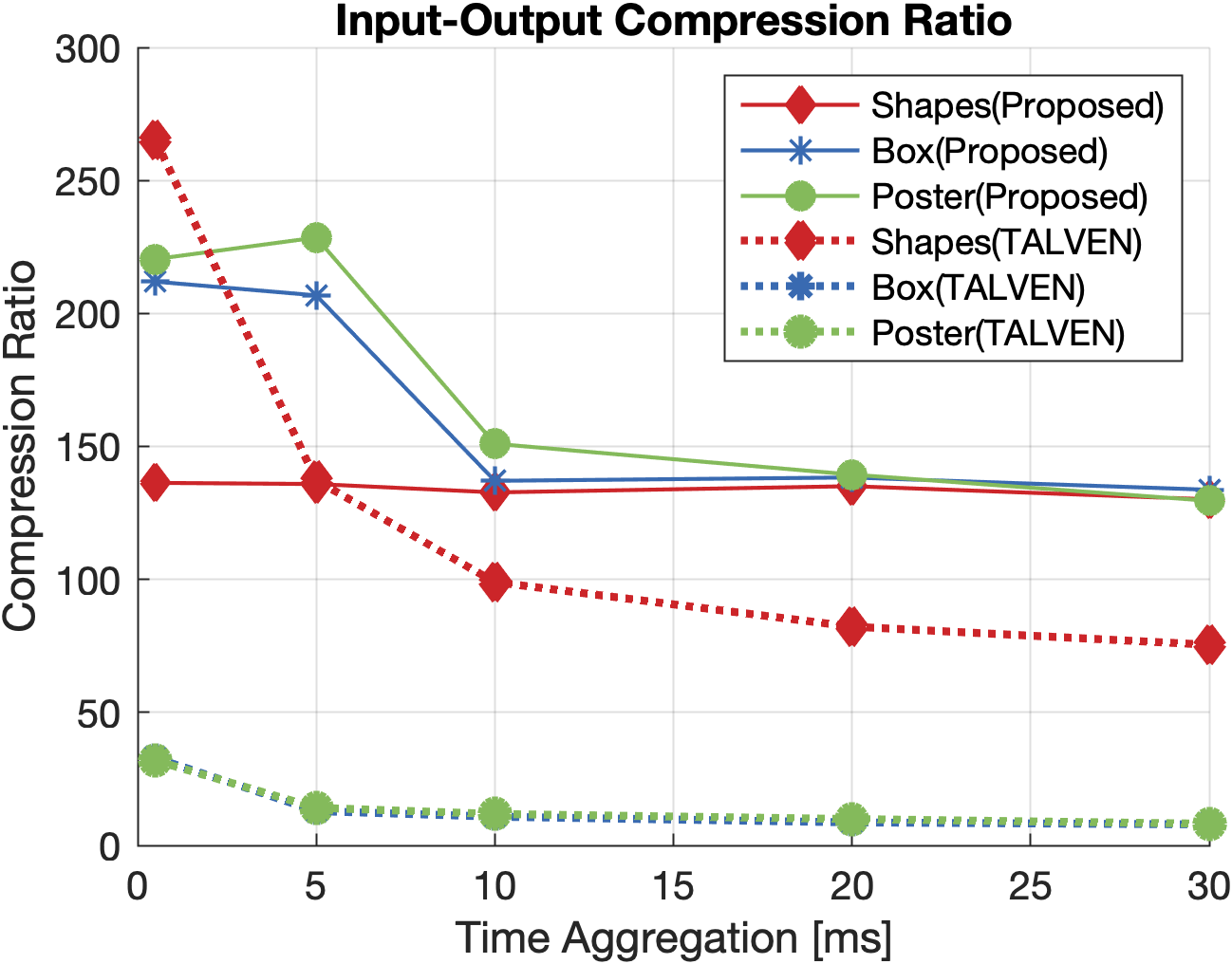}}
    \caption{}
    \label{IO_CR}
     \end{subfigure}
     \begin{subfigure}{0.31\textwidth}
    \centerline{\includegraphics[width=\textwidth]{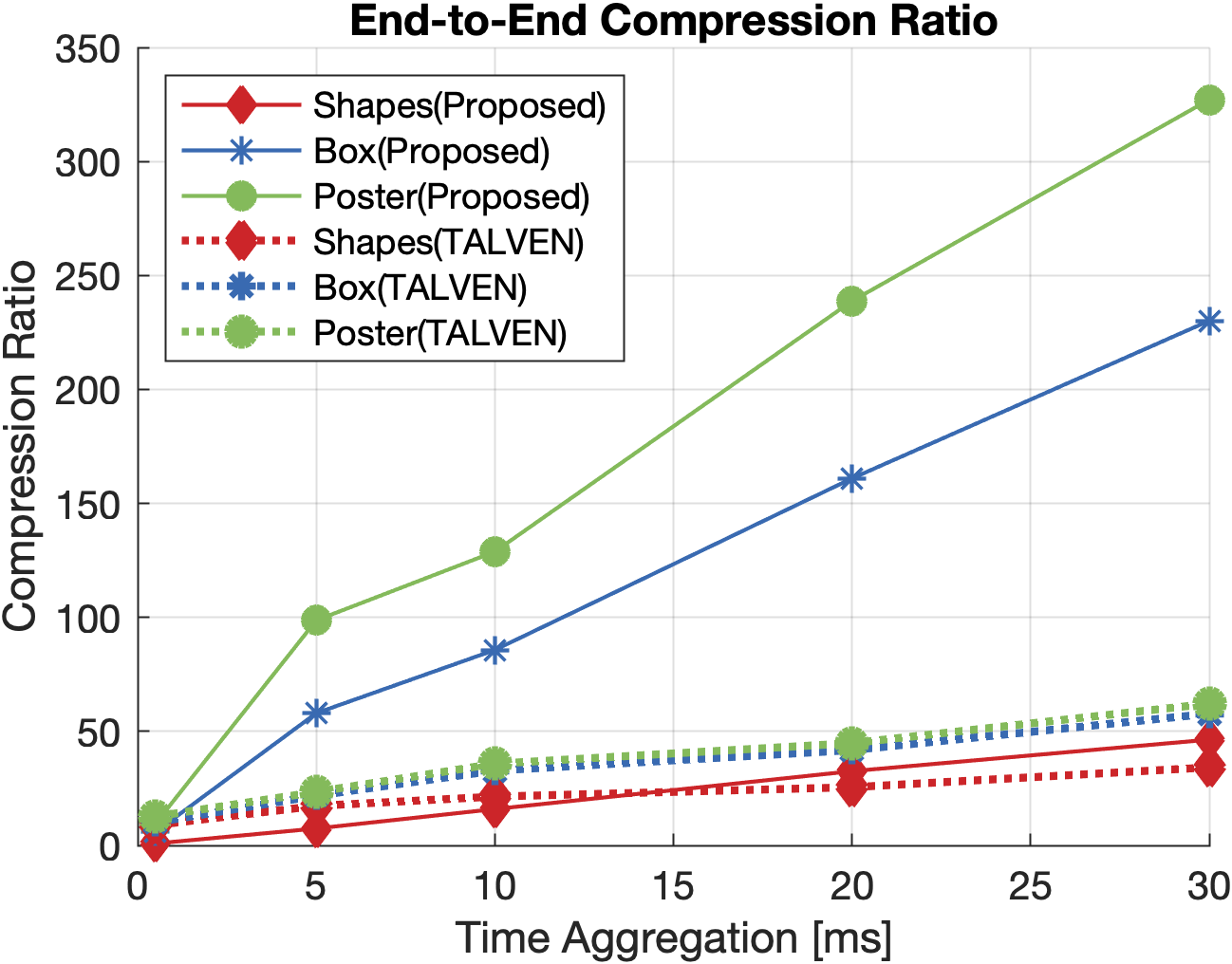}}
    \caption{}
    \label{E2E_CR}
     \end{subfigure}
     \begin{subfigure}{0.31\textwidth}
    \centerline{\includegraphics[width=\textwidth]{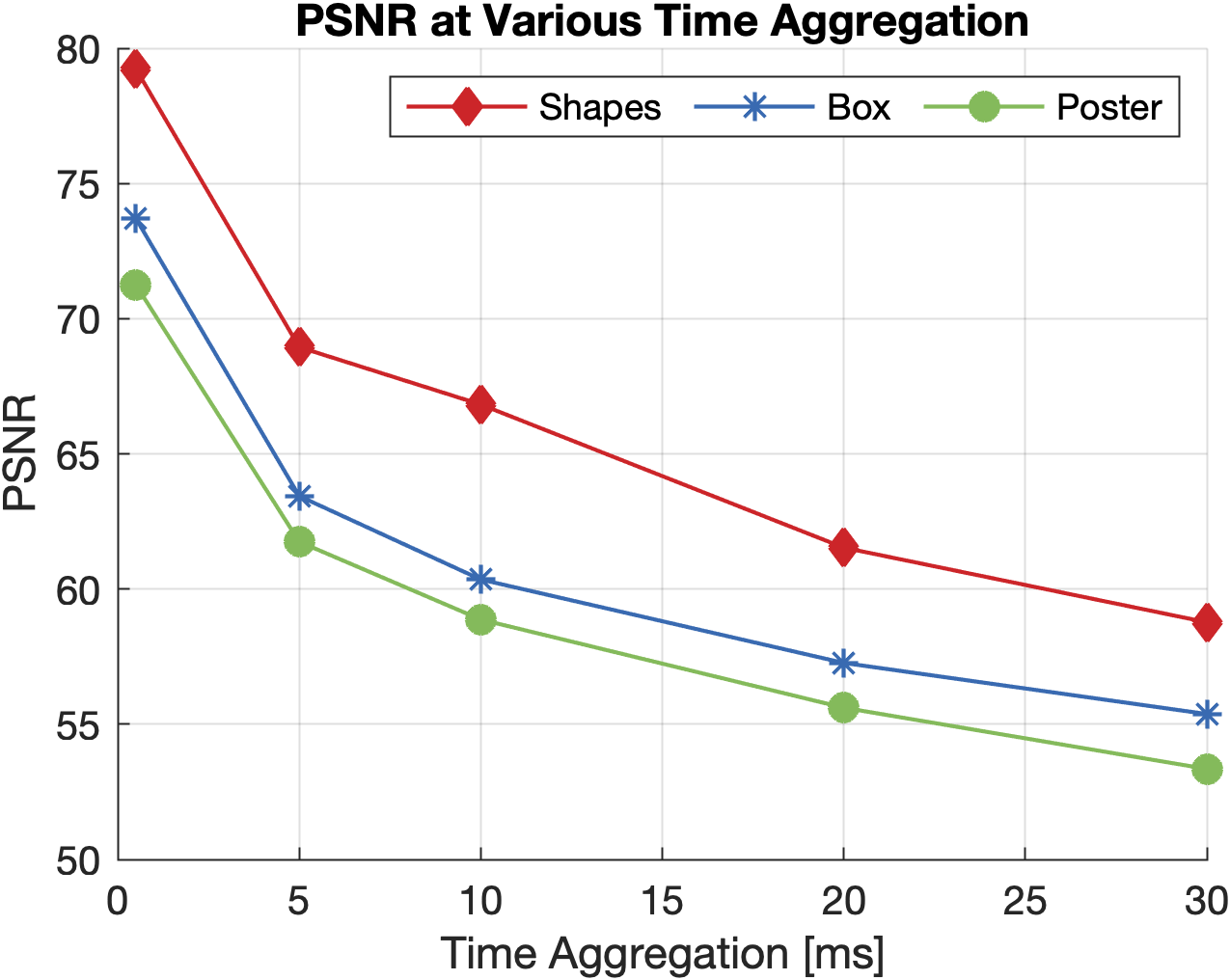}}
    \caption{}
    \label{psnr}
     \end{subfigure}
    \caption{Comparison of compression performance of Proposed Scheme with respect to TALVEN. (a) Input-Output Compression Ratio (b) End-to-end Compression Ratio (c) PSNR at various Time Aggregation.}
\end{figure}
\subsection{Implementation Details}
To test our compression scheme's performance, we used standard event camera datasets~\cite{mueggler2017event}. We choose three different datasets \textit{Boxes}, \textit{Poster} and \textit{Shapes} to evaluate the competence of our proposed scheme at varying degree of scene complexity and camera movement speed. The evaluation was done on a workstation with Intel Xeon 2.30GHz CPU, Nvidia K80 12GB GPU and 12GB RAM specifications. All the experiments were conducted using the Pytorch framework.

The auto-encoder framework comprises of an encoder with layers size (30 $\times$ 30)-1000-500-250-20 and a symmetric decoder. The units in the code layer are linear, while other units are logistics. The model's accuracy increases with lesser units in the code layer, but it has the drawback of longer training time and an increase in low-dimensional code blocks. Hence, we settled with a 20-dimensional code layer maintaining a balance between training time, accuracy and the number of code blocks. The 20-dimensional code blocks are compressed using an entropy-based Huffman coding technique generating a compressed DVS bitstream. The network was trained on 96,000 event blocks derived from the first 10 seconds of the event sequence with 10ms time-aggregration for 20 epochs and 10 batch size. The average training time for each sample was around 30 minutes, with the learning rate set at 0.1. The network was evaluated with test samples obtained with various degree of time-aggregation (0.5ms, 5ms, 10ms, 20ms and 30ms).

\subsection{Experimental Results}
\label{experimental_results}
As shown in Table~\ref{tab_CR}, we compared our performance with existing benchmark strategies regarding end-to-end compression ratio on standard datasets with diverse scene complexity and camera movement speed. We outperform all the benchmark coders at almost all time-aggregation values. We also compared our performance with a state-of-the-art event data coder proposed by Khan et al.~\cite{khan2020time} in terms of end-to-end compression and input-output compression ratio. In both cases of compression metrics, our proposed compression scheme outperforms the TALVEN method~\cite{khan2020time} at all time-aggregation values as depicted in Figure~\ref{E2E_CR} and  Figure~\ref{IO_CR} respectively. In addition, higher event rate sequences produce better compression performance than low event rate sequences. The \textit{Poster} sequence, for example, has a very high event rate (4.01 Mega-events/s), while the \textit{Shapes} sequence has the lowest event rate (0.245 Mega-events/s). The \textit{Shapes} sequence has low scene complexity and camera speed. Naturally, it should have high compression gains while, in reality, the \textit{Poster} sequence yields high compression gain. In sequences with high event rate, the delta-coded timestamp need lesser bits to be encoded when compared to low event rate sequences. Additionally, we computed the reconstruction quality of our proposed model with the help of the peak-signal-to-noise ratio (PSNR) metric. The proposed model can reconstruct the dynamic vision sensor event data from the compressed bitstream with a high PSNR value, as shown in Figure~\ref{psnr}.  

% Table generated by Excel2LaTeX from sheet 'Sheet6 (2)'
\begin{table}[!t]
  \caption{Comparison of End-to-End Compression Ratio of Proposed Scheme with Benchmark Coders.}
    \centering
    \resizebox{0.5\textwidth}{!}{
    \begin{tabular}{|c|c|c|c|c|}
    \hline
          & \multicolumn{1}{|c|}{\textbf{Time Aggregation}} & \multicolumn{1}{|c|}{\textbf{Shapes}} & \multicolumn{1}{|c|}{\textbf{Boxes}} & \multicolumn{1}{|c|}{\textbf{Poster}} \\
    \hline
    \multicolumn{1}{|c|}{\multirow{5}{*}{\begin{sideways}\textbf{Proposed}\end{sideways}}} & \textbf{0.5ms} & 0.837 & 5.885 & 9.516 \\
\cline{2-2}          & \textbf{5ms} & 7.406 & 57.983 & 98.697 \\
\cline{2-2}          & \textbf{10ms} & 15.935 & 85.715 & 128.834 \\
\cline{2-2}          & \textbf{20ms} & 32.554 & 160.974 & 238.774 \\
\cline{2-2}          & \textbf{30ms} & 46.381 & 230.125 & 327.093 \\
    \hline
    \hline
    \multirow{8}{*}{\begin{sideways}\textbf{Benchmark}\end{sideways}} & \textbf{SPIKE Coding} & 3.78  & 4.95  & 4.88 \\
\cline{2-5}          & \textbf{Huffman} & 1.79  & 1.96  & 1.79 \\
\cline{2-5}          & \textbf{Zstd} & 2.67  & 4.13  & 4.02 \\
\cline{2-5}          & \textbf{LZMA} & 3.04  & 4.92  & 4.77 \\
\cline{2-5}          & \textbf{LZ4} & 2.19  & 3.03  & 2.97 \\
\cline{2-5}          & \textbf{ZLib} & 2.8   & 4.21  & 4.12 \\
\cline{2-5}          & \textbf{Brotli} & 2.78  & 4.38  & 4.26 \\
\cline{2-5}          & \textbf{Snappy} & 2.21  & 2.98  & 2.92 \\
    \hline
    \end{tabular}%
    }
  \label{tab_CR}%
\end{table}%
\section{Conclusion}
Dynamic vision sensor captures the change in per pixel intensity value producing an asynchronous stream of event data. We proposed a novel framework for event data arrangement and its compression based on DBN. It is among the first few to utilise a deep learning framework for event data processing. Time aggregation of spike events is advantageous because it enhances efficiency by bringing a specific group of events into context with one another. The super-frame arrangement yields high spatial and temporal correlations among events which the DBN utilises to obtain latent feature code. The proposed coding framework outperforms the benchmark compression schemes as supported by the extensive performance comparison study presented in the paper.

\bibliographystyle{unsrt}  
\bibliography{root}  %%% Remove comment to use the external .bib file (using bibtex).

\begin{thebibliography}{10}

\bibitem{mahowald1994silicon}
Misha Mahowald.
\newblock The silicon retina.
\newblock In {\em An Analog VLSI System for Stereoscopic Vision}, pages 4--65.
  Springer, 1994.

\bibitem{lichtsteiner2008128}
Patrick Lichtsteiner, Christoph Posch, and Tobi Delbruck.
\newblock A $128\times128$ $120$ db $15\mu s$ latency asynchronous temporal
  contrast vision sensor.
\newblock {\em IEEE journal of solid-state circuits}, 43(2):566--576, 2008.

\bibitem{brandli2014240}
Christian Brandli, Raphael Berner, Minhao Yang, Shih-Chii Liu, and Tobi
  Delbruck.
\newblock A 240$\times$ 180 130 db 3 $\mu$s latency global shutter
  spatiotemporal vision sensor.
\newblock {\em IEEE Journal of Solid-State Circuits}, 49(10):2333--2341, 2014.

\bibitem{posch2010live}
Christoph Posch, Daniel Matolin, Rainer Wohlgenannt, Michael Hofst{\"a}tter,
  Peter Sch{\"o}n, Martin Litzenberger, Daniel Bauer, and Heinrich Garn.
\newblock Live demonstration: Atis camera with full-custom ae processor.
\newblock In {\em Proceedings of IEEE ISCAS}, pages 1392--1392. IEEE, 2010.

\bibitem{kim2008simultaneous}
Hanme Kim, Ankur Handa, Ryad Benosman, Sio-Hoi Ieng, and Andrew~J Davison.
\newblock Simultaneous mosaicing and tracking with an event camera.
\newblock {\em J. Solid State Circ}, 43:566--576, 2008.

\bibitem{vidal2018ultimate}
Antoni~Rosinol Vidal, Henri Rebecq, Timo Horstschaefer, and Davide Scaramuzza.
\newblock Ultimate slam? combining events, images, and imu for robust visual
  slam in hdr and high-speed scenarios.
\newblock {\em IEEE Robotics and Automation Letters}, 3(2):994--1001, 2018.

\bibitem{rebecq2019events}
Henri Rebecq, Ren{\'e} Ranftl, Vladlen Koltun, and Davide Scaramuzza.
\newblock Events-to-video: Bringing modern computer vision to event cameras.
\newblock In {\em Proceedings of the IEEE/CVF CVPR}, pages 3857--3866, 2019.

\bibitem{zhu2019unsupervised}
Alex~Zihao Zhu, Liangzhe Yuan, Kenneth Chaney, and Kostas Daniilidis.
\newblock Unsupervised event-based learning of optical flow, depth, and
  egomotion.
\newblock In {\em Proceedings of the IEEE/CVF CVPR}, pages 989--997, 2019.

\bibitem{chan2007aer}
Vincent Chan, Shih-Chii Liu, and Andr van Schaik.
\newblock Aer ear: A matched silicon cochlea pair with address event
  representation interface.
\newblock {\em IEEE TCAS-1}, 54(1):48--59, 2007.

\bibitem{bi2018spike}
Zhichao Bi, Siwei Dong, Yonghong Tian, and Tiejun Huang.
\newblock Spike coding for dynamic vision sensors.
\newblock In {\em 2018 Data Compression Conference}, pages 117--126. IEEE,
  2018.

\bibitem{collet2018zstandard}
Yann Collet and Murray Kucherawy.
\newblock Zstandard compression and the application/zstd media type.
\newblock Technical report, 2018.

\bibitem{deutsch1996zlib}
Peter Deutsch and Jean-Loup Gailly.
\newblock Zlib compressed data format specification version 3.3.
\newblock Technical report, 1996.

\bibitem{alakuijala2016brotli}
Jyrki Alakuijala and Zoltan Szabadka.
\newblock Brotli compressed data format.
\newblock Technical report, 2016.

\bibitem{blalock2018sprintz}
Davis Blalock, Samuel Madden, and John Guttag.
\newblock Sprintz: Time series compression for the internet of things.
\newblock {\em Proceedings of the ACM on Interactive, Mobile, Wearable and
  Ubiquitous Technologies}, 2(3):1--23, 2018.

\bibitem{lemire2015decoding}
Daniel Lemire and Leonid Boytsov.
\newblock Decoding billions of integers per second through vectorization.
\newblock {\em Software: Practice and Experience}, 45(1):1--29, 2015.

\bibitem{maqueda2018event}
Ana~I Maqueda, Antonio Loquercio, Guillermo Gallego, Narciso Garc{\'\i}a, and
  Davide Scaramuzza.
\newblock Event-based vision meets deep learning on steering prediction for
  self-driving cars.
\newblock In {\em Proceedings of the IEEE/CVF CVPR}, pages 5419--5427, 2018.

\bibitem{rigi2018novel}
Amin Rigi, Fariborz Baghaei~Naeini, Dimitrios Makris, and Yahya Zweiri.
\newblock A novel event-based incipient slip detection using dynamic
  active-pixel vision sensor (davis).
\newblock {\em Sensors}, 18(2):333, 2018.

\bibitem{mueggler2014event}
Elias Mueggler, Basil Huber, and Davide Scaramuzza.
\newblock Event-based, 6-dof pose tracking for high-speed maneuvers.
\newblock In {\em 2014 IEEE/RSJ International Conference on Intelligent Robots
  and Systems}, pages 2761--2768. IEEE, 2014.

\bibitem{lagorce2016hots}
Xavier Lagorce, Garrick Orchard, Francesco Galluppi, Bertram~E Shi, and Ryad~B
  Benosman.
\newblock Hots: a hierarchy of event-based time-surfaces for pattern
  recognition.
\newblock {\em IEEE TPAMI}, 39(7):1346--1359, 2016.

\bibitem{mueggler2017event}
Elias Mueggler, Henri Rebecq, Guillermo Gallego, Tobi Delbruck, and Davide
  Scaramuzza.
\newblock The event-camera dataset and simulator: Event-based data for pose
  estimation, visual odometry, and slam.
\newblock {\em The International Journal of Robotics Research}, 36(2):142--149,
  2017.

\bibitem{khan2020time}
Nabeel Khan, Khurram Iqbal, and Maria~G Martini.
\newblock Time-aggregation-based lossless video encoding for neuromorphic
  vision sensor data.
\newblock {\em IEEE Internet of Things Journal}, 8(1):596--609, 2020.

\end{thebibliography}

\end{document}